\documentclass[manuscript]{acmart}
\AtBeginDocument{%
  }

\setcopyright{acmlicensed}
\copyrightyear{2018}
\acmYear{2018}
\acmDOI{XXXXXXX.XXXXXXX}
\acmConference[FAccT '27]{ACM Conference on Fairness, Accountability, and Transparency }{June 03--05,
  2018}{Woodstock, NY}
\acmISBN{978-1-4503-XXXX-X/20/06}

\usepackage{listings}
\usepackage[utf8]{inputenc}
\usepackage{textgreek}
\usepackage{xcolor}

\usepackage{enumitem}

\usepackage{booktabs}
\usepackage{tabularx}
\usepackage{array}
\usepackage{makecell}
\usepackage{threeparttable}

\floatstyle{plaintop}
\restylefloat{table}
\usepackage{xcolor}
\usepackage[table]{xcolor}

\usepackage{booktabs}
\usepackage{array}
\usepackage{threeparttable}
\newsavebox{\tabbox}
\usepackage{fvextra}
\usepackage{orcidlink}

\begin{document}

\title{Honest Brokers in AI Regulation: How Public Comments Introduce Evidence to Regulators}

\author{Sarah Tran\,\orcidlink{0009-0004-1816-5062}}
\affiliation{%
  \institution{University of Washington Information School}
  \city{Seattle}
  \country{Washington}}
\email{saraht45@uw.edu}

\author{Anna-Maria Gueorguieva\,\orcidlink{0009-0006-4850-9318}}
\affiliation{%
  \institution{University of Washington Information School}
  \city{Seattle}
  \country{Washington}
}

\author{Eva Maxfield Brown\,\orcidlink{0000-0003-2564-0373}}
\affiliation{%
  \institution{University of Washington Information School}
  \city{Seattle}
  \country{Washington}
}

\author{Nicholas Weber\,\orcidlink{0000-0001-5592-2510}}
\affiliation{%
  \institution{University of Washington Information School}
  \city{Seattle}
  \country{Washington}
}


\renewcommand{\shortauthors}{Tran et al.}

\begin{abstract} The regulation of Artificial Intelligence is susceptible to various forms of capture by industry actors that hold undue influence over the policymaking process, due in part to their information advantage in developing these technologies. In the following paper, we investigate the roles of public commenters in providing evidence to potential regulation in the USA. We build a corpus of 40 potential regulations, and 2,664 public comments submitted by a variety of stakeholders. We computationally process these texts to extract evidence meant to inform regulators, in the form of formal citations and informal references. We find significant differences in the type and amount of evidence supplied: Holding comment length constant, business commenters introduce altogether less evidence, and government, university, and civil society commenters provide significantly more research and legal sources. The publication status of cited research is nearly evenly split, with preprints and peer-reviewed articles being introduced by commenters of all types. We further explore patterns of self-citation and find that while it is prevalent amongst all groups, a small cluster of organizations account for over half of all self-citations in our corpus. Our findings complicate a narrative that public comments alone are means to information capture. We interpret these results in terms of the role that commenters attempt to play in the policymaking process---being either an Issue Advocate or an Honest Broker. These findings on the supply of evidence in public comments shed light on an often overlooked form of policymaking that has implications for how and through what sources information is made accessible to regulators of AI.    

\end{abstract}



\received{20 February 2007}
\received[revised]{12 March 2009}
\received[accepted]{5 June 2009}

\maketitle

\section{Introduction}

What counts as credible evidence, and who supplies it, is a recurring point of contention in U.S. policymaking \cite{furnas2025partisan}. In 2021, the Biden administration issued a memorandum that committed federal agencies to “make evidence-based decisions guided by the best available science and data” \cite{biden2021memorandum}. Four years later, the Trump administration withdrew the memorandum \cite{eo14154} and issued Executive Order 14303, “Restoring Gold Standard Science,” which directs agencies to ground their decisions in science that is reproducible and transparent \cite{eo14303}. Although issued by two administrations with sharply different views on the role of government regulation, these directives reflect a common logic in the executive branch: evidence should be the foundation of policymaking. But, they also raise substantive questions about which kinds of science and evidence are seen as credible and legitimate, and how federal agencies should be informed in policy development.

Controversies around evidence and credible science are particularly salient for AI governance. Regulating AI is complicated by a number of factors, including the technical capabilities of systems that advance faster than available evidence of their impacts \cite{bengio2025international}, AI’s growing share of worldwide economic growth \cite{rubinton2026tracking}, the absence of a single body positioned to form expert consensus \cite{bommasani2025neurips}, and the need to address public safety \cite{casper2025pitfalls, bommasani2025advancing}. While U.S. regulations formally govern firms operating in the United States, the risks those rules address cross national borders, and international efforts to assess AI safety reflect this global need \cite{bengio2025international}. 

In the following study, we investigate how public commenters use evidence in their attempts to advocate for or against proposed U.S. regulations on artificial intelligence. To do so, we build a corpus of 40 dockets (e.g., proposed regulations or requests for information) that received a total of 2,664 substantive comments. We process these texts for computational analysis, and then we extract formal citations and informal references, and link these sources to canonical metadata about the publication venue and authors. We ask:\\ 
\indent \textbf{RQ 1} What evidence do public comments supply when responding to potential regulation?\\
\indent \textbf{RQ 2 }How do organizational affiliations align with the evidence that is cited?\\
Our goal is to provide a quantitative description of the evidence introduced into federal AI policymaking. Prior work identifies AI regulation as vulnerable to capture \cite{wei2024how, birhane2026big}, including information capture, in which regulated parties use the volume and technical complexity of information to gain control over regulatory decisions \cite{wagner2010filter}. Drawing on the history of the tobacco and fossil fuel industries \cite{michaels2006manufactured, reed2021playbook, casper2025pitfalls}, this work argues that AI firms pursue comparable strategies: concentrating control over research resources such as data and compute \cite{ahmed2020dedemocratization, bakcoleman2025industry}, funding academic research, and sponsoring academic events \cite{abdalla2021grey}. These accounts address how industry shapes the production of research. We examine the evidence that reaches regulators through formal policy procedures. We classify the types of evidence that public commenters cite, the commenters’ relationship to that evidence, and analyze the cited science by its publication status. We focus part of our analysis on scientific evidence because regulators give data and research particular weight when evaluating comments \cite{topazian2024understanding}. 

We find little support that public comments \textit{alone} are a means to information capture. In our data, business/industry commenters do not supply more technical information and do not cite formal literature at higher rates than other commenters. We do find that commenters supply significantly different amounts and types of evidence to policymakers, with university, government, and civil society commenters supplying more research and legal citations. We interpret these differences in citation behavior by adapting science policy's typology of roles that experts take when informing decisions marked by high value conflict and high uncertainty \cite{pielke2007honest}: 1. \textit{Issue Advocates} align themselves with a particular policy and introduce evidence that narrows the potential solutions based on their own experience or understanding; 2. \textit{Honest Brokers }focus on broadening the potential solutions to policy by introducing evidence that represents diverse views and emerging scientific understanding. Viewed through this typology, we interpret differences in the type and amount of evidence introduced by public commenters as means to either broaden or narrow policy choices.   

\section{Background}
In the United States, federal power is divided among three branches: Congress makes laws, the executive branch implements and enforces those laws, and the courts interpret the fair and just application of laws. Congress has long delegated much of its lawmaking authority to federal agencies, which turn broad statutes into specific rules\footnote{In the U.S., ``rule'' and ``regulation'' mean the same thing: a binding requirement issued by a federal agency to carry out a law passed by Congress. Throughout, we use ``rule'' and ``rulemaking'' interchangeably with ``regulation'' and ``regulations.''} \cite{yackee2019politics}. Under the Administrative Procedure Act, agencies generally create rules through a notice-and-comment procedure: an agency publishes a proposed rule, invites public comment via written submission online or public hearings, and must consider the comments it receives before issuing a final rule \cite{apa553}. Scholars have linked congressional gridlock in the legislative branch to the presidents’ growing reliance on unilateral tools, such as executive orders and agency rulemaking, to achieve their administrative agenda \cite{lowande2021presidential}. This erosion of checks and balances, a hallmark of the federal system of democracy in the U.S., makes rulemaking procedures, like public commenting, an important domain to understand how regulation is shaped and codified by the executive branch \cite{alanoca2025comparing}.

Scholars have long examined how organized interests shape the rule-making process, describing forms of influence that range from routine persuasion to regulatory capture \cite{yackee2006sweet, carpenter2013preventing}. In legislation, influence is most visible in lobbying, where firms and interest groups spend heavily to gain access to congress and shape potential laws \cite{drutman2015business}. Rulemaking differs critically in that it includes a formal, public channel through which anyone can submit written comments on a proposed rule. Compelled by law, most federal agencies post proposed rules, requests for information, and the comments they receive to regulations.gov, a public docket system that is openly accessible \cite{gsa2025regulationsgov}. This procedural and bureaucratic information environment of rulemaking provides the empirical backbone of our study---we use public comments to understand how evidence is introduced and used for shaping regulation. 

\subsubsection{Influence of Public Comments.}
A large body of political science and public administration research examines whether public commenters influence agency rules, which groups are most influential, and under what conditions. Early studies found little evidence that business participation gives industry disproportionate influence \cite{golden1998interest, nixon2002friends}. Subsequent work that has benefited from access to more data about federal rulemaking has expanded the scope of these early findings on influence. Yackee and colleagues show that agencies do respond to both the number and the content of comments, and well-resourced commenters, particularly businesses, have disproportionate influence on final rules \cite{yackee2006sweet, yackee2006bias, nelson2012coalitions}. Regulators’ own accounts of the public comment process also emphasize that the substance of a comment matters to the actions they subsequently decide to take. In a survey of 115 former EPA regulators, Topazian et al. found that respondents valued comments containing data and technical recommendations \cite{topazian2024understanding}. 

More recent work examines what agency responsiveness looks like in practice. Love analyzes 854 federal rules issued between 2017 and 2023 and finds that roughly two-thirds of agency citations to public comments are not accompanied by any change to the binding regulatory text \cite{love2026hearing}. Agencies often cite comments they ultimately reject, particularly those from well-resourced groups. Love argues that agencies acknowledge comments to build legally defensible records while preserving their policy autonomy. These competing findings reflect the complexity of federal rulemaking. Theories of bureaucratic policymaking describe regulators as arbiters who balance the demands of competing interests \cite{peltzman1976toward}, and as decision makers who weigh information supplied by interested parties whose goals may differ from their own \cite{schnakenberg2023formal}. We contribute to this work by examining how different commenters supply evidence in federal AI policymaking, a domain marked by technical, social, political, and economic complexity and uncertainty.

\subsubsection{Capture.}
A concern parallel to influence in rulemaking is regulatory capture, in which regulated entities use their position to shape rules that benefit them or preserve their power \cite{li2023regulatory}. Capture is a particular concern for AI regulation because regulators rely heavily on technical expertise that regulated firms are often best positioned to produce \cite{wei2024how, bakcoleman2025industry}. Prior work suggests that capture of AI policy can operate through the production and framing of evidence \cite{birhane2026big}. Similarly, Yew and Judge show that technical claims about AI safety can redirect oversight toward voluntary standards that favor AI firms \cite{yew2025antiregulatory}. Wagner describes a more general form of this dynamic as “information capture,” in which the volume of technical information that well-resourced participants submit gives them an advantage in rulemaking \cite{wagner2010filter}. While these arguments are compelling accounts of how capture might operate in policy generally, the influence of public comments is difficult to trace empirically. As Carpenter argues, alignment between a rule and one group’s preferences does not by itself establish capture \cite{carpenter2013detecting}, and studies that take a computational approach to analyzing large volumes of comments find at best weak associations between what commenters request and what final rules contain \cite{balla2022flood, fan2026who}. These works characterize participants and their positions using text-as-data methods such as topic modeling \cite{karakanta2026whose}, preference coding \cite{tallberg2023ai}, narrative analysis \cite{stewart2026narratives}, and quote-grounded extraction \cite{bertaglia2026traceable}. A second line of work examines the evidence itself and shows that its type and source can track the interests of those who supply it. Hemmerich et al. find that opponents of FDA tobacco rules relied more on non-peer-reviewed and conflicted sources than supporters did \cite{hemmerich2017evidentiary}. Outside public comment, Furnas et al. show that congressional committees and think tanks aligned with different parties cite largely different scientific literatures, with only ~5\% of cited papers drawing citations from both sides \cite{furnas2025partisan}. These findings are important to our work, because if evidence sources sort by political actor, they may also sort by commenter type in AI rulemaking. We acknowledge that evidence in public comments takes forms other than citation, as commenters may offer personal narrative and moral reasoning to support their positions \cite{stewart2019giving, fabi2022public}. We build on this work by linking the type, source, and status of evidence in AI regulation comments to commenter type, a link that we argue brings information capture closer to direct observation.

\section{Empirical Approach}
In the following section we describe the sources of our data, and how we analyze this information to make defensible claims about the supply of evidence to policymakers. A practical obstacle for this work is that requests for comments on AI regulation draw thousands of submissions, and the most substantive of these are provided as PDF attachments. The form of a comment, as a fixed PDF document with numerous formatting standards, makes extraction and claim-level coding difficult \cite{bertaglia2026traceable, karakanta2026whose}. In this paper, we computationally extract evidence types and sources from PDF comment submissions and link each piece of evidence to the commenter who introduced it. This approach makes it possible to examine information capture at a formative stage of rulemaking, as a distribution of cited evidence across commenter populations, rather than inferring it from regulatory outcomes \cite{wagner2010filter, fan2026who}.

\subsection{Data Collection}
We collect all federal dockets that may be relevant to artificial intelligence regulation by querying the \href{https://api.govinfo.gov/docs/}{GovInfo’s Federal Register API}. We include in our corpus only dockets that are  (1) “strongly related” to AI and (2) result in a government response, such as guidance documents, whitepapers, and reports. We consider a docket to be “strongly related” to AI if it mentions AI or related terms (e.g., open weight model or large language model). We include only those that received a response as this signifies a serious attempt by a federal agency to engage with the supplied evidence of the public comments. To do so, we manually review the narrative preamble of each docket, known as the Request for Information (RFI) section. This resulted in 40 dockets across 10 federal agencies related to our study (see Appendix~\ref{sec:app_data_collection} for details on data collection).

We retrieve metadata for each docket (e.g., name of agency issuing the RFI and the number of comments received) using the \href{https://www.federalregister.gov/developers/documentation/api/v1}{Federal Register API}, or by bulk downloading comments posted to \href{https://www.nitrd.gov/}{NITRD}. For comments that were not made publicly available for a docket ($n$ = 146) we submitted a FOIA request to the associated federal agency. This process returned a total of 13,698 comments. 

Public comments can be submitted as a free response to a webform or an attached PDF. For comments submitted as PDF, we remove all comment attachments that are less than 2 pages, as prior work has identified that it is unlikely that these comments contain substantive content \cite{costa2019public}. This resulted in the exclusion of 10,452 comments, 8,327 (79.67\%) of which were submitted by individuals, and among the remaining, 2,125 (20.33\%) were roughly only a paragraph long (mean = 183.7 words; median = 113 words). We further exclude three categories of comment attachments. First, we remove files with little or no directly extractable text (e.g., .xlsx, .jpeg, and .png files). Second, we remove supplementary information, such as appended news articles or reports, attached to a comment because our focus is on the evidence commenters cite directly when appealing to regulators. Finally, we remove comment submissions from mass-comment campaigns, identified as exact or near-duplicate comments submissions within a docket, since prior work finds that mass comments have little measurable influence on final rules \cite{balla2022flood}. The final corpus contains 2,664 comments. See Appendix~\ref{sec:app_filtering_public_comments} for further details on comment exclusion and Appendix~\ref{sec:app_comment_deduplication} for comment de-duplication method. 

\subsection{Public Comment Data Processing}

\subsubsection{Public Comment Text Pre-Processing.} 
We convert all comments stored in PDFs (n = 2,664) into plain text or markdown using \texttt{MarkerV2}. We first remove formatting and header information, strip whitespace, and convert special characters. We then remove spurious information that might bias our extraction of evidence, such as email/mailto links, and in-line links that reference the Request for Information or Executive Order being responded to. 


\begin{table}[htbp]
\centering
\small
\caption{Classification performance by label.}
\label{tab:classification-performance}
\begin{tabular}{lccc}
\toprule
\textbf{Label} & \textbf{Precision} & \textbf{Recall} & \textbf{F1} \\
\midrule
\texttt{citation\_sentence}           & 0.819 & 0.896 & 0.856 \\
\texttt{citation\_link}               & 0.800 & 0.822 & 0.811 \\
\texttt{citation\_footnote/reference} & 0.810 & 0.811 & 0.810 \\
\texttt{author\_name}                 & 0.894 & 0.876 & 0.885 \\
\texttt{author\_affiliation}          & 0.758 & 0.649 & 0.699 \\
\bottomrule
\end{tabular}
\end{table}

\subsubsection{Extraction of Citation Sentences.}
Text spans that include a reference or citation are the main unit of analysis in our study. We first annotate a sample of 6 comments to identify these spans. Using inter-rater reliability (IRR) across four raters on 153 citation text spans, we evaluate our ability to correctly identify the span location and categorize each evidence type and commenter (e.g., citation\_sentence, citation\_link, citation\_footnote, citation\_reference, author, or author\_affiliation). We achieve substantial agreement on span location (κ = 0.71) and perfect agreement on span labels (κ = 1.00). We then individually annotate another 101 documents to develop a ground-truth dataset containing 4,813 total citation text spans, consisting of 2,000 citation sentences, 1,356 citation links, and 1,457 citation footnotes and references. 

Prior work shows that few-shot prompting performs well on citation context tasks~\cite{koloveas2025can}. We adapt this approach to extract citations from every public comment in our corpus. In an initial annotation, we found that comments cite evidence in three ways: hyperlinks, footnotes, and reference lists. We developed a separate few-shot prompt for each citation style using \texttt{gemini-2.5-flash}. Each prompt extracts three features from a comment’s plain text: (1) sentence-citation pairs, consisting of the citing sentence and the evidence it cites; (2) the author’s name and affiliation; and (3) the reference list, if present. We refined each prompt iteratively and manually reviewed the outputs. The final prompts used were evaluated against an annotated test dataset. See Appendix~\ref{sec:app_prompts} for the prompts used, and Appendix~\ref{sec:app_model_eval} for additional details on evaluation and results. 

Next, we evaluate the model’s citation extraction – including citation sentences, citation links, and citation footnotes/references – against our ground truth dataset. We recognize that span overlap is the standard metric for evaluating text extraction and segmentation, but because the model does not return span locations, we instead follow \cite{breton2025leveraging} and evaluate extractions using two string-matching criteria: (1) exact matches and (2) near matches, defined as normalized Levenshtein similarity of at least 0.85. In Table~\ref{tab:classification-performance} we report the results of the prompting strategy against our ground truth dataset, achieving F1 scores that match our distance measure threshold. Finally, we process the entire corpus, extracting citation spans, using this approach. See Appendix~\ref{sec:app_model_eval} for the full procedure.

\subsubsection{Classifying Commenter Type.}
 We categorize public commenters into six types based on their organizational affiliation: individual, civil society organization, trade/coalition organization, business/industry, university, and government. We obtain the commenter name and organizational affiliations from the \href{https://open.gsa.gov/api/regulationsgov/}{Regulations.gov API} metadata. Where these fields are missing, we extract this information directly from the comment text. For organizational commenters, we programmatically assign a North American Industry Classification System (NAICS) code to determine which type of organization the authors belong to (NAICS is the federal standard for classifying organizations by their primary economic activity). See Appendix~\ref{sec:app_commenter_type} for details on this procedure. 

\begin{figure}[htbp]
    \centering
    \includegraphics[width=1.0\textwidth]{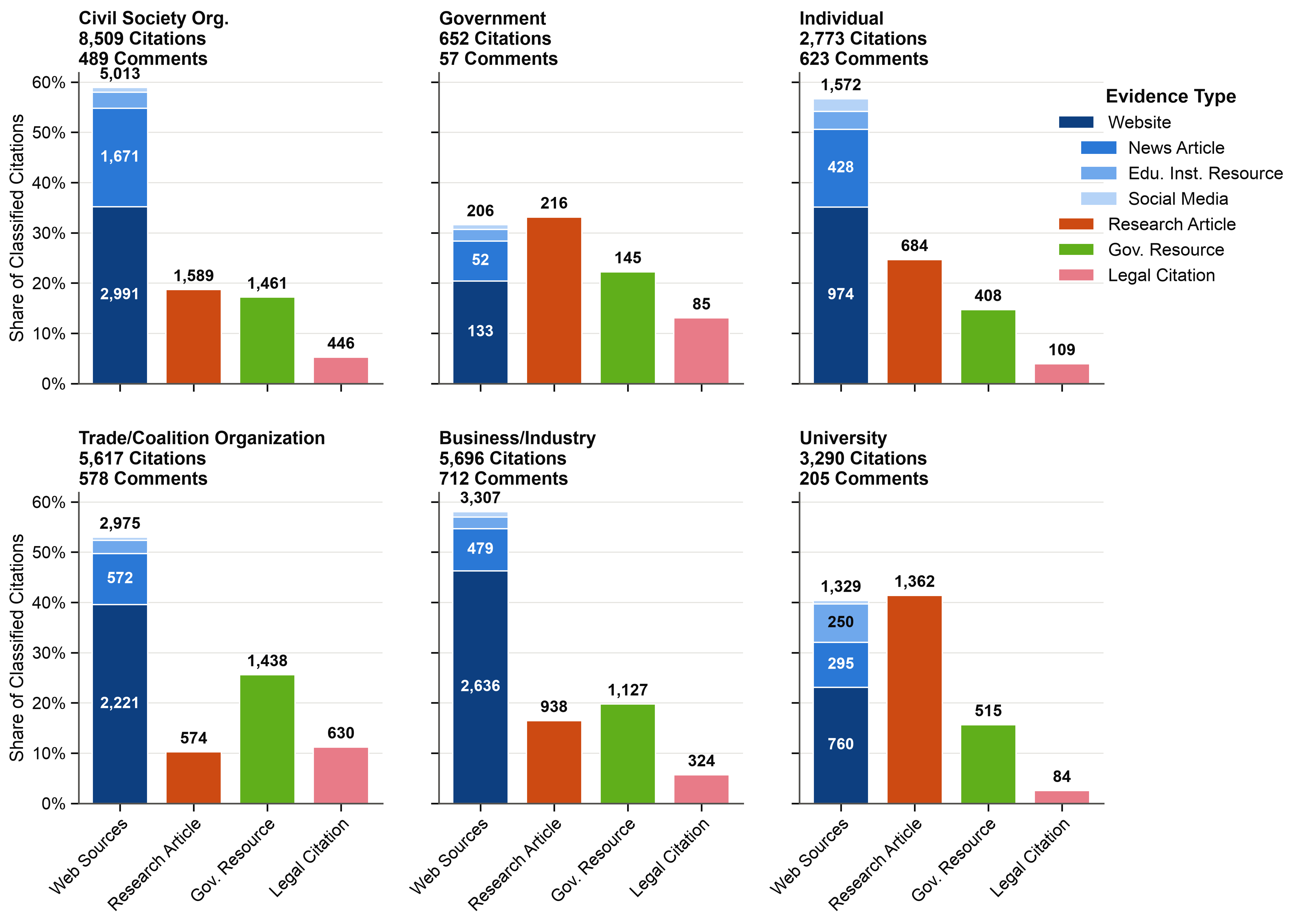}
    \caption{Distribution of evidence cited, by commenter type}
    \label{fig:evidence_type_shares}
\end{figure}

\subsubsection{Classifying Evidence Type.} 
Following prior work \cite{costa2019public}, we classify each cited source into one of four evidence categories: research article, legal citation, government resource, or website. We further divide websites into news articles, educational resources, and social media posts. We classified citations in two steps. First, we applied rule-based detection (e.g., pattern matching on legal citation formats and URL domains such as \texttt{.gov}). To identify research articles, we also queried Crossref and OpenAlex with each reference's text to check whether it matched an indexed academic work. See Appendix~\ref{sec:app_bib_data} for rule-based detection method and post-processing steps for each evidence type. Second, we resolved the remaining unclassified citations with a retrieval-assisted classification step. For each citation, Claude Opus 5 searched the web using the citation text, identified the cited source, assigned it to an evidence category from our codebook, and returned structured metadata (URL and domain) with a self-reported confidence rating. We retained only high-confidence classifications, 1,355 of 2,714 citations (49.93\%), and coded the remainder manually. To validate this step, we manually checked a random sample of 623 retained classifications (46\% of retained), verifying both the assigned category and the returned metadata against the source. Of these, 620 (99.52\%) were correct. See Appendix~\ref{sec:app_evi_type} for further details on this procedure, and breakdown of the remaining unclassified citations. 

\subsubsection{Retrieving Bibliographic Data of Cited Research Articles.}
We retrieve bibliographic metadata for every cited research article in two steps. First, we resolve each citation to a persistent identifier. When the citation link or reference text contains a DOI or arXiv ID, we extract it directly ($n$=2,504 citations). Otherwise, we query the \href{https://www.crossref.org/documentation/retrieve-metadata/rest-api/}{Crossref API} with the reference text provided by a commenter. For each query, the Crossref API provides a rule-based confidence score based on title similarity, publication year, and first-author agreement. We accept matches scoring 0.90 or higher ($n$=770 citations). The remaining citations to research articles are parsed for title, author(s), venue, and year.
Second, we retrieve metadata from \href{https://openalex.org/}{OpenAlex}, \href{https://www.semanticscholar.org/}{Semantic Scholar}, \href{https://arxiv.org/}{arXiv}, \href{https://www.crossref.org/}{Crossref}, and \href{https://dblp.org/}{DBLP}, querying by identifier for resolved citations and by parsed fields for the rest. Each source has gaps in its coverage of disciplines, author affiliations, and preprint versions, so combining them gives more complete records than any single source. When sources disagree, we prefer Semantic Scholar. We also record each article’s publication and preprint status as of the date the comment was submitted. Appendix~\ref{sec:app_bib_data} describes the matching procedure and its validation. 

\subsection{Methods}


Our research questions concern the types of evidence public commenters supply and how it varies with the organizations they represent. We operationalize these questions by modeling how a commenter's organization affiliation relates to four outcomes: the amount of evidence cited, the types of evidence cited, the publication status of cited research, and whether commenters cite evidence their own organization produced. We first present the specification common to all models, then the features specific to each analysis.

\subsubsection{Shared Specifications.}
All models use \textit{Commenter Type} as the primary predictor, with ``business/industry'' as the reference level. We chose business/industry because it is the largest group in our dataset ($n = $ 709) and because our primary interest is in how other commenters’ citation practices compare to this organization type. Next, all models control for comment length, measured as the log of the word count and centered at the mean so that the intercept represents a comment of typical length. We log-transform comment length to reduce skew and to account for its proportional relationship with citation counts. We measure comment length as the word count of each comment's cleaned text, which does not contain citation references or links (see Appendix~\ref{sec:app_clean_md} for procedure). Moreover, all models include random intercepts for \textit{Commenter Organization} and \textit{Docket}. Organizations frequently comment on multiple dockets, so comments from the same organization are not independent; dockets also differ in topic and in the questions they pose, which induces clustering among comments within a docket.

\subsubsection{Evidence Use by Commenter Type.}
\label{sec:Evidence_use_by_Commenter_Type}
We examine whether a commenter's organizational affiliation is related to both how much they cite and the type of evidence that they supply. To do so, we first model the relationship between \textit{Commenter Type} and the \textit{number of citations per comment} to test for systematic differences in the quantity of evidence each commenter type introduces. We use a negative binomial regression with the shared specification above. We also model the relationship between \textit{Commenter Type} and the \textit{Evidence Type} each commenter type cites, again using a negative binomial regression. The predictors are \textit{Commenter Type}, \textit{Evidence Type}, and their interaction. We set websites as the reference level for \textit{Evidence Type} because it is the largest category (Figure~\ref{fig:evidence_type_shares}). Because the data are in long format, with one row per comment and evidence type, we include a random intercept for comment to account for repeated observations of the same comment. We then compare each commenter type against business/industry within each evidence type, using Dunnett’s adjustment for multiple comparisons.

\subsubsection{Peer-Reviewed vs.\ Preprints.}
\label{sec:peer-reviewed_vs_preprints}
We compare how often each group cites peer-reviewed articles versus preprints to understand how public commenters navigate the emergent, fast-paced nature of AI research in their policy advocacy. For this analysis, we use the publication status of the research article at the time the comment was submitted. For example, an article cited as a preprint that was later peer-reviewed is coded as a preprint. Then, among comments that cite at least one research article, we model the number of preprints cited out of the total number of research articles cited, using a beta-binomial generalized linear mixed model with the shared specifications. We compare all pairs of commenter types using estimated marginal means with Tukey’s adjustment. See Appendix~\ref{sec:app_bib_data} for further details.

\subsubsection{Self-Citation.}
\label{sec:self-citation}
We test whether commenters favor evidence produced by their own organization. We code a citation as a self-citation when the cited work has an author affiliated with the commenter’s organization. For example, an OpenAI-affiliated commenter citing a paper with an OpenAI-affiliated author counts. When a comment lists several authors or affiliations, a match with any of them qualifies. We match organizations by canonical name. See Appendix~\ref{sec:app_commenter_type}; Appendix~\ref{sec:ident_self_cit} gives the full procedure for identifying self-citations.

We model two outcomes. The first is whether a comment contains any self-citation, estimated with mixed-effects logistic regression. The second is the \textit{count of self-citations per comment}, estimated with negative binomial regression. Both models use the shared specification, which controls for comment length and includes a random intercept for organization. Both also add the log of total citations as a covariate, since comments with more citations have more opportunities to self-cite. Finally, we use each organization’s predicted random intercept from the negative binomial model to identify organizations whose self-citation count exceeds what is typical for their commenter type, holding comment length and citation count constant.


\section{Results}

\subsection{Variance in commenter quantity and type of evidence}
\begin{figure}[htbp]
    \centering
    \includegraphics[width=1.0\textwidth]{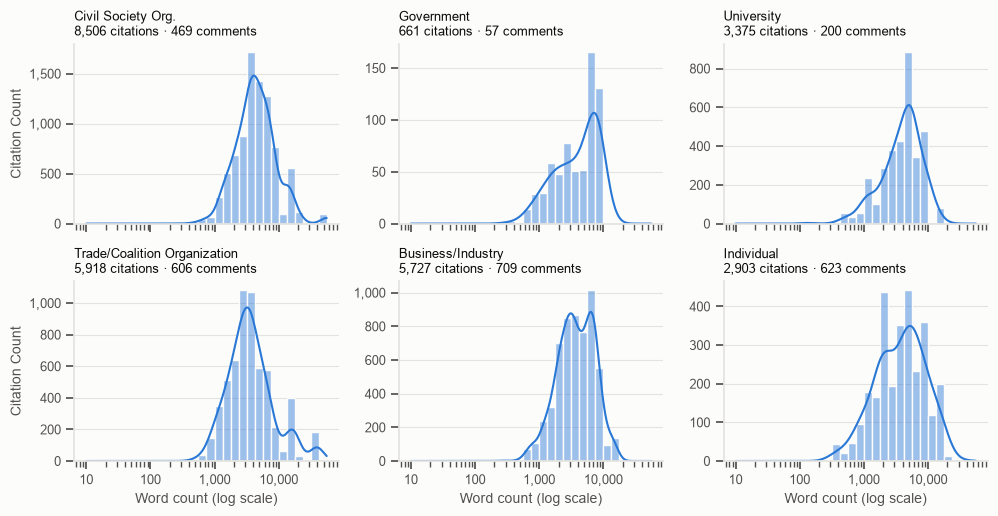} 
    \caption{Citation Density by Commenter Type}
    \label{fig:citation_density}
\end{figure}

Commenter types differ in both the quantity and the type of evidence they cite. Holding comment length constant, government, civil society, trade/coalition, and university commenters included 1.7 to 2.4 times as many citations as business/industry commenters (Table~\ref{tab:citation_density}; Appendix Figure~\ref{fig:citation_density}). Pairwise comparisons (Table~\ref{tab:pairwise-citations-count}) sort commenters into three tiers based on citation amount: individuals cited the fewest sources, business/industry commenters cited more, and government, civil society, trade/coalition, and university commenters cited the most, with no significant differences among the four in the final tier. Comment length was also strongly associated with citation count (IRR = 3.53 per unit increase in log word count). This association reflects the tendency for longer comments to cite more evidence. These results show that citation volume is not only a function of how much a commenter writes. At equal length, a comment from a university or civil society organization carries more than twice the documented evidence of a comment from a business. This suggests that commenter types write in different evidentiary registers: some ground their claims in external sources, while others rely more on assertion or on the authority of the commenter itself.

\begin{figure}[H]
    \centering
    \includegraphics[width=\columnwidth]{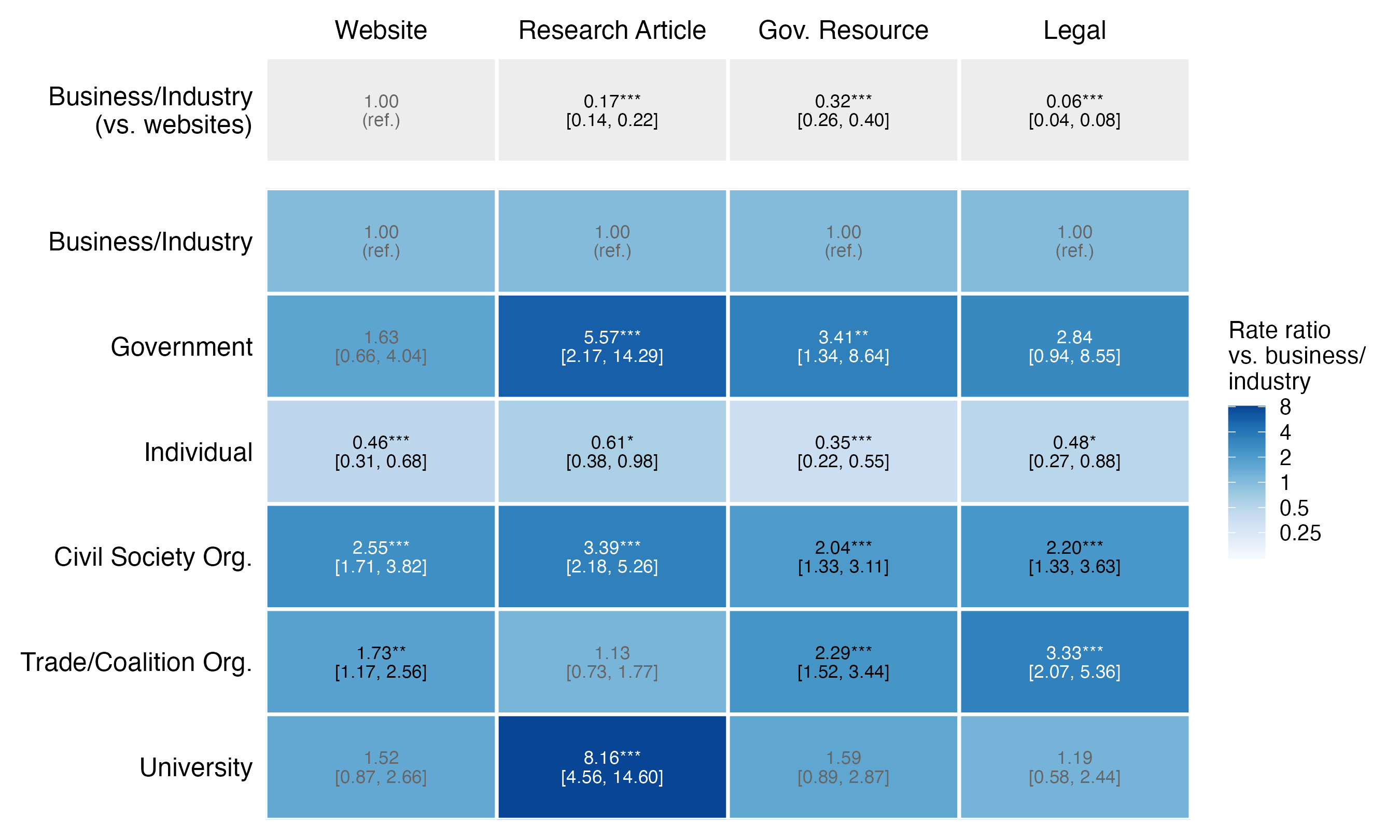}
    \caption{Rate ratios of expected evidence type cited relative to business/industry commenters}
    \label{fig:rq1_heatmap}
\end{figure}

Commenter types also differ in the evidence types they cite. Figure~\ref{fig:rq1_heatmap} reports, for each evidence type, the expected citation count for each commenter type relative to business/industry. Because commenter types differ in total citations, these ratios combine differences in overall volume with differences in emphasis. We therefore read a group’s preference for an evidence type as how far its ratio for that type departs from its ratios for other types (and not from the size of any single ratio alone).

Within business/industry comments, it is striking that the expected counts of cited research articles, government resources, and legal sources were 83\%, 68\%, and 94\% lower, respectively, than their expected count of websites. Against this baseline, two commenter types show distinct patterns. University commenters cited research articles at 7.84 times the rate of business/industry commenters, well above their overall citation ratio of 2.40. Their advantage is concentrated in research rather than spread across evidence types. Government commenters cited government resources at 3.33 times the rate of business/industry commenters, again above their overall ratio of 2.15. Each group’s distinctive evidence corresponds to the kind of knowledge its institution produces: universities cite research, and government agencies cite other government sources.

Civil society organizations depart from this pattern, citing every evidence type more often than business/industry commenters, at 2.04 to 3.47 times the rate. This range underscores their overall ratio of 2.32, which indicates that their higher citation counts reflect \textit{volume} rather than a preference for any one evidence type. Civil society commenters appear to act as generalists, drawing on whatever sources support their position. Individuals cited fewer sources than business/industry commenters across all evidence types. This is consistent with prior work showing that individual commenters tend to submit low-information comments~\cite{dinour2017potato}. Trade/coalition organizations cited websites, government resources, and legal sources at higher rates than business/industry commenters (and did not differ significantly on research articles). This divergence matters because trade associations typically represent industry interests. Their citation practices suggest a division of labor: firms submit comments that rely on informal sources, while the associations representing them engage regulators with legal and governmental evidence. We return to this finding in Section~\ref{sec:discussion}. 

\begin{table*}[t]
\centering
\small
\captionsetup{justification=centering, singlelinecheck=true, margin=0pt}
\caption{Mixed-effects negative binomial regression of total citations per comment}
\label{tab:citation_density}
\sbox{\tabbox}{%
\begin{tabular}{lrrrcrl}
\toprule
Predictor & $b$ & SE & IRR & 95\% CI & \% change & $p$ \\
\midrule
Intercept (Business/Industry) & 1.389 & 0.108 & 4.01 & [3.24, 4.96] & --- & $<$ .001 \\
\quad Government & 0.767 & 0.261 & 2.15 & [1.29, 3.59] & $+$115.3 & .003 \\
\quad Individual & $-$0.479 & 0.127 & 0.62 & [0.48, 0.79] & $-$38.0 & $<$ .001 \\
\quad Civil Society Org. & 0.842 & 0.122 & 2.32 & [1.83, 2.95] & $+$132.1 & $<$ .001 \\
\quad Trade/Coalition Org. & 0.510 & 0.115 & 1.67 & [1.33, 2.08] & $+$66.5 & $<$ .001 \\
\quad University & 0.876 & 0.165 & 2.40 & [1.74, 3.32] & $+$140.1 & $<$ .001 \\
Log word count (centered) & 1.261 & 0.046 & 3.53 & [3.23, 3.86] & $+$252.8 & $<$ .001 \\
\bottomrule
\end{tabular}}
\usebox{\tabbox}\par
\vspace{4pt}
\end{table*}


\begin{table}[t]
\centering
\small
\caption{Mixed-effects logistic regression of self-citation, by commenter type}
\label{tab:logit-commenter}
\begin{tabular}{lrrrrcr}
\toprule
Predictor & $b$ & SE & $z$ & OR & 95\% CI & $p$ \\
\midrule
Intercept (\textit{Business/Industry}) & 0.041 & 0.172 & 0.24 & 1.04 & [0.74, 1.46] & .812 \\
\rowcolor{gray!30}
Government                    & 1.024 & 0.479 & 2.14 & $\times$2.78 & [1.09, 7.12] & .033 \\
Civil Society Org.            & 0.234 & 0.224 & 1.05 & $\times$1.26 & [0.82, 1.96] & .295 \\
Trade/Coalition Org.          & 0.371 & 0.215 & 1.73 & $\times$1.45 & [0.95, 2.21] & .084 \\
\rowcolor{gray!30}
University                    & 0.695 & 0.305 & 2.28 & $\times$2.00 & [1.10, 3.64] & .023 \\
\bottomrule
\end{tabular}
\end{table}

\subsection{Self-citation is concentrated among a small group of commenters}
In the previous analysis we looked at the supply of evidence by public commenters by type. A parallel question is whether commenters tend to rely on evidence they created themselves (such as their own papers, websites, or blog posts). We characterize this as a form of self-citation---where an author of a public comment refers to content that was produced by their own organization. Table~\ref{tab:logit-commenter} demonstrates that commenter type is associated with whether a comment contained any self-citation. Relative to business/industry, comments from government and university commenters had 2.78 and 2.00 times the odds of containing a self-citation, respectively. However, no pairwise difference between commenter types remained significant after Tukey adjustment. Business/industry commenters are therefore not distinctive in their self-citation. About half of business/industry comments contain a self-citation, but this does not differ reliably from other commenter types. Commenter type was also not associated with the number of self-citations per comment among comments with at least one self-citation (Appendix Table~\ref{tab:nb_self_citation}). 

\begin{table*}[t]
\centering
\small
\caption{Pairwise comparisons of expected citation counts by commenter type}
\label{tab:pairwise-citations-count}
\begin{tabular}{lrrrrcr}
\toprule
Contrast & $b$ & SE & $z$ & Ratio & 95\% CI & $p$ \\
\midrule
\quad Business/industry vs. & & & & & & \\
\rowcolor{gray!30}
\qquad Government                 & $-0.766$ & 0.260 & $-2.94$ & 0.47 & [0.22, 0.98] & .039 \\
\rowcolor{gray!30}
\qquad Individual                 & 0.479    & 0.126 & 3.78    & 1.61 & [1.13, 2.31] & .002 \\
\rowcolor{gray!30}
\qquad Civil society organization & $-0.842$ & 0.122 & $-6.89$ & 0.43 & [0.30, 0.61] & $<$.001 \\
\rowcolor{gray!30}
\qquad Trade/coalition            & $-0.509$ & 0.114 & $-4.45$ & 0.60 & [0.43, 0.83] & $<$.001 \\
\rowcolor{gray!30}
\qquad University                 & $-0.877$ & 0.165 & $-5.31$ & 0.42 & [0.26, 0.67] & $<$.001 \\
\quad Government vs. & & & & & & \\
\rowcolor{gray!30}
\qquad Individual                 & 1.245    & 0.279 & 4.47    & 3.47 & [1.57, 7.69] & $<$.001 \\
\qquad Civil society organization & $-0.075$ & 0.263 & $-0.29$ & 0.93 & [0.44, 1.96] & $>$.999 \\
\qquad Trade/coalition            & 0.257    & 0.261 & 0.98    & 1.29 & [0.61, 2.72] & .924 \\
\qquad University                 & $-0.109$ & 0.290 & $-0.38$ & 0.90 & [0.39, 2.05] & .999 \\
\quad Individual vs. & & & & & & \\
\rowcolor{gray!30}
\qquad Civil society organization & $-1.321$ & 0.150 & $-8.82$ & 0.27 & [0.17, 0.41] & $<$.001 \\
\rowcolor{gray!30}
\qquad Trade/coalition            & $-0.989$ & 0.139 & $-7.12$ & 0.37 & [0.25, 0.55] & $<$.001 \\
\rowcolor{gray!30}
\qquad University                 & $-1.355$ & 0.177 & $-7.66$ & 0.26 & [0.16, 0.43] & $<$.001 \\
\quad Civil society organization vs. & & & & & & \\
\qquad Trade/coalition            & 0.332    & 0.124 & 2.67    & 1.39 & [0.98, 1.99] & .080 \\
\qquad University                 & $-0.034$ & 0.174 & $-0.20$ & 0.97 & [0.59, 1.59] & $>$.999 \\
\quad Trade/coalition vs. & & & & & & \\
\qquad University                 & $-0.365$ & 0.170 & $-2.16$ & 0.69 & [0.43, 1.13] & .258 \\
\bottomrule
\end{tabular}
\end{table*}

Although these differences are not statistically significant, self-citation \textit{is} a common practice, and is concentrated among a small number of organizations: 55.8\% of the 779 commenters self-cited at least once, but the top 5\% of commenters (n = 38) accounted for 54.8\% ($n$ = 2,093 of 3,817) of all self-citations. Of these 38 commenters, 13 (34.21\%) are civil society organizations, 10 (26.32\%) are business/industry organizations, 9 (23.68\%) are universities, and 6 (15.79\%) are trade/coalition organizations. The estimated standard deviation of the organization-level random intercept was $\sigma = 0.90$ (95\% CI [0.77, 1.04]) on the log scale, implying that the middle 95\% of organizations self-cite at between 0.17 and 5.85 times the rate of a typical organization of the same commenter type. We note that our self-citation measure depends on affiliation metadata, which is more complete for research articles than for websites and other grey literature. Commenter types that cite more research, such as universities and government, may have their self-citations detected more completely, which would inflate their estimates relative to business/industry.

Our interest in self-citation is motivated by Lauber et al.’s analysis of public health comments, which found that industry commenters referred to their own evidence more often than other commenters \cite{lauber2021commercial}. We therefore expected business/industry commenters to self-cite more often than other commenter types. Instead, self-citation is not specific to any one commenter type; it is concentrated among particular organizations within each type. 

\subsection{Public comments cite peer-reviewed articles and preprints at similar rates}

\begin{figure}[t]
    \centering
    \includegraphics[width=\columnwidth]{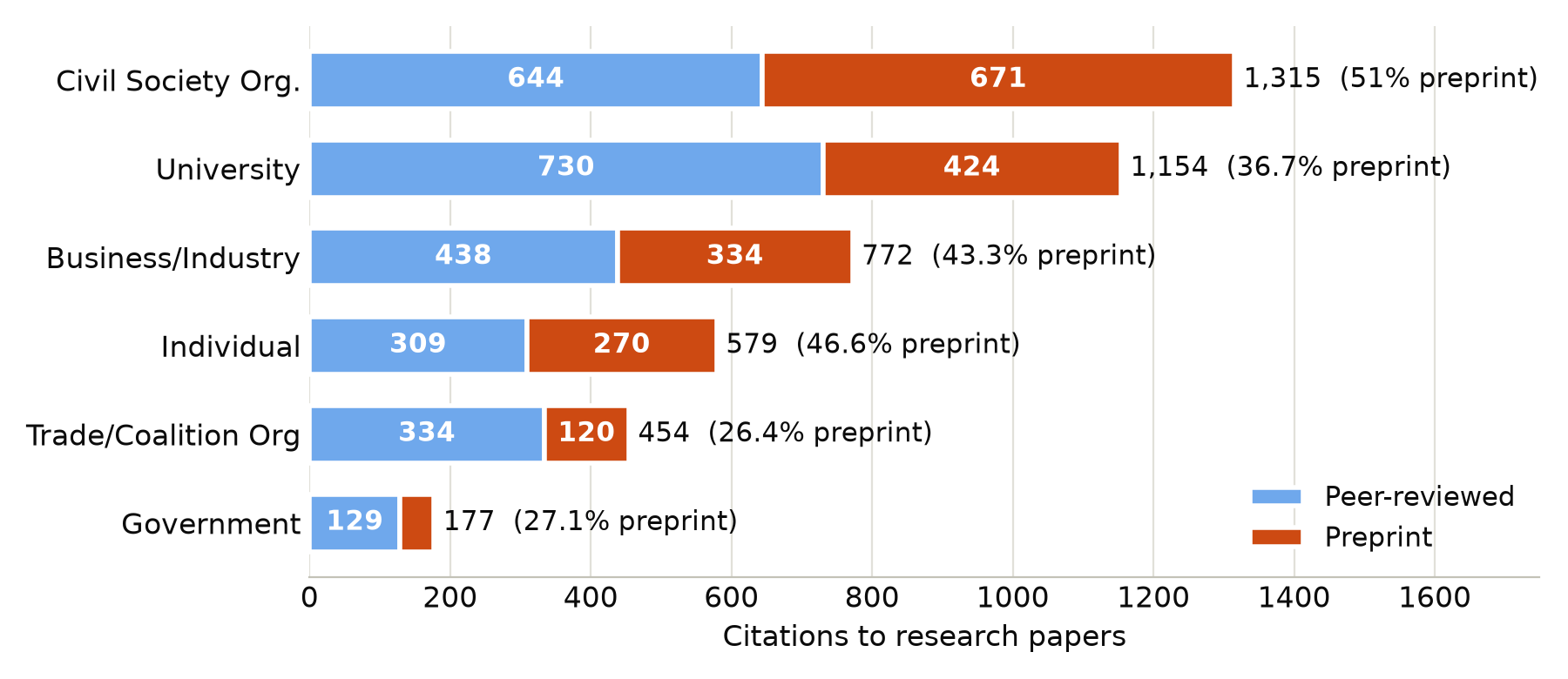}
    \caption{Citation of Peer-reviewed Research vs Preprints}
    \label{fig: peer-reviewed}
\end{figure}

Thus far, we have examined the types of evidence commenters introduce and how they relate to organizational affiliation. We next examine where in the research lifecycle cited AI research sits. We classify each cited research article as peer-reviewed or as a preprint on the date the comment was submitted. We do not treat peer review as a proxy for quality; we use it to measure how early AI research enters the policymaking process.

Preprints accounted for 41.9\% of citations to research articles (1,867 of 4,451; Figure~\ref{fig: peer-reviewed}). Raw preprint shares ranged from 26.4\% among trade/coalition organizations to 51.0\% among civil society organizations. However, commenter type was not associated with the share of cited research articles that were preprints once we accounted for comment length and for clustering within organizations (Table~\ref{tab:betabinom-peer-reviewed}). No commenter type differed from business/industry, and no pairwise comparison was significant after Tukey adjustment. The raw differences therefore reflect variation among individual organizations more than systematic differences between commenter types. Counting each preprint once, 401 of the 921 distinct preprints that were cited (43.5\%) later went on to be published in a peer-reviewed venue. This is lower than the rate of up to 66\% reported for computer science preprints~\cite{sutton2017popularity}, but this gap likely reflects a lag in publication since recently cited preprints have had less time to be published.

Our results indicate that citing preprints is common practice across commenter types in AI regulation. Two features of AI research may explain this. First, preprints are a routine part of how AI and computer science research is disseminated, and practitioners widely accept them~\cite{zhou2025everyone}. Second, preprints are open access, while some peer-reviewed work requires a journal subscription. 

\subsection{Self-citation and preprints by commenter type}
As an exploratory follow-up, we check whether public commenters that engage in self-citation also cite research preprints at different rates. Table~\ref{tab:selfcite} reports these patterns by commenter type. In the previous section, we found that universities self-cited most often, in 40.3\% of works, compared with 13.7\% to 15.6\% for business/industry, government, and civil society commenters. Trade/coalition organizations rarely self-cited (1.6\%). However, when combining these two analyses we see a different pattern emerge: Business/industry commenters directed 64.1\% of their self-citations to preprints (95\% CI: 55.1, 72.2), compared with 48.6\% for civil society organizations (41.4, 55.9) and 38.5\% for universities (34.2, 43.0).  Universities self-cited more often, but they mostly cited published work. Business/industry commenters self-cited less often, and when they did, they mostly cited work that had not undergone peer review. We report this pattern as a direction for further study.

\begin{table}[ht]
\centering
\small
\begin{threeparttable}
\caption{Self-citation and preprint self-citation by commenter type}
\label{tab:selfcite}
\begin{tabular}{lrrrrl}
\toprule
Commenter type & Works & \makecell[r]{Self-\\citations} & \makecell[r]{Self-citation\\rate} & \makecell[r]{Preprint\\self-citations} & \makecell[l]{\% to preprints\\(95\% CI)} \\
\midrule
Business/industry            & 748   & 117 & 15.6\% & 75  & 64.1\% (55.1, 72.2) \\
Government                   & 173   & 27  & 15.6\% & 5   & 18.5\% (8.2, 36.7) \\
Civil Society                    & 1,302 & 179 & 13.7\% & 87  & 48.6\% (41.4, 55.9) \\
Trade/coalition organization & 443   & 7   & 1.6\%  & 2   & 28.6\% (8.2, 64.1) \\
University                   & 1,146 & 462 & 40.3\% & 178 & 38.5\% (34.2, 43.0) \\
\bottomrule
\end{tabular}
\end{threeparttable}
\end{table}

\section{Discussion}
\label{sec:discussion}
The regulation of AI is one of the most difficult challenges in a complicated technology policy landscape. As a result, the task is often compared to nuclear energy \cite{hatz2025nuclear} in terms of its complexity and safety implications, and to tobacco in terms of its potential for information capture \cite{bakcoleman2025industry, abdalla2021grey}. Our work provides a quantitative description of the supply of evidence to federal policymakers as they shape AI regulation in the USA. We find that there is a significant difference between both the quantity and type of evidence that commenters cite. Industry commenters cite mostly informal sources (websites) and, apart from low-information individual commenters, cite significantly less of all types of evidence. We find that university and government commenters are the most likely to cite the scientific literature, with the latter citing laws and government resources at greater than three times business/industry commenters. This complicates a narrative that, in public comments, business entities are attempting to capture regulators with an information advantage. Carpenter has argued that alignment between a rule and one group’s preferences does not by itself establish capture in public commenting \cite{carpenter2013detecting}. But in AI regulation specifically, industry commenters also have an enormous advantage over these other groups in influencing policy through the media \cite{birhane2026big}, lobbying \cite{cheng2024ailobbyists}, and congressional testimony \cite{leach2026shared}. Our findings about disparities in evidence citations are nonetheless meaningful---they represent what AI industry actors are willing to put down on paper and submit for public consideration. That these entities cite significantly less evidence overall conflicts deeply with their own calls for ``evidence-based policymaking'' that are, at least in this domain, a potential delay tactic to advancing meaningful regulations \cite{casper2025pitfalls}. 

Our work also reflects the fact that public comments may be a unique outlet for experts to inform policymakers based on their position in industry, academia, or civil society---as self-citation is a frequent occurrence (over half of all commenters). However, we also observe a small group of actors—not isolated in one type of organization—engaging in significantly higher rates of self-citation; just 38 commenters accounted for ~55\% of all self-citations in our corpus. When combining our analysis of self-citation and preprints we see that Business commenters most often refer to their own research. This finding may speak to a form of information capture attempted by actors who are predominantly introducing evidence of their own making. 

We also investigate the role that emerging evidence plays in informing AI regulation by looking at the publication status of cited research articles. We find that 41.9\% of all research citations are to preprints, and of those 43.5\% went on to be peer-reviewed. The rate of preprint citation tracks nearly identically with Suh and Byrne's finding that the citation of preprints in AI policy documents (not public comments) rose from 5\% in 2015 to 40\% in 2024~\cite{suh2025shifting}. We interpret this as a function of the evidence base of AI relevant to policymaking---public commenters seek to provide regulators with the most up-to-date, and the most relevant emerging evidence regardless of its publication status. 

We introduced Pielke’s typology, which distinguishes experts by whether they expand or narrow the set of policy options available to decision makers~\cite{pielke2007honest}. Our data do not record which options a comment endorses, but they do record three properties of the evidence each public comment supplies: its volume across evidence types, its reliance on the commenter’s own work, and its status in the research lifecycle. We treat these as indicators of a comment’s capacity to broaden or narrow the evidence base before regulators, not as direct measures of a commenter’s role.
At the level of commenter types, these three indicators separate commenters more clearly than a simple contrast between business/industry and all other organizations. Civil society organizations cited every evidence type at 2 to 3 times the rate of business/industry commenters, with no single type dominating. These commenters are acting as \textit{honest brokers} in the policy process, by introducing evidence from several sources that leaves a range of options open. University and government commenters also cited more evidence, but concentrated on the kind of evidence their institutions produce. Universities cited research articles, and government commenters cited government resources. This specialization is closer to Pielke’s description of a \textit{science arbiter}, who answers questions of policy choice by supplying factual accounts they are best positioned to supply. Business/industry commenters cited less evidence of every type and relied mostly on websites. This is clear evidence that these organizations are playing a role of \textit{issue advocate} by submitting comments that rely more on status and organizational experience than on an external evidence base. Similarly, trade/coalition organizations cited legal and government sources at higher rates than the firms they often represent, which suggests issue advocacy conducted in the legal and regulatory terms agencies use. 

However, our self-citation results suggest that these roles might belong to organizations more than to sectors. Variation in self-citation across organizations was larger than any difference between commenter types, and the 38 organizations responsible for most self-citations came from every type except government and individuals. A comment that cites heavily but draws mainly on its author’s own work narrows the evidence base it presents regardless of its author's affiliation. Under this reading, the risk of information capture through public comments lies with a small set of organizations that repeatedly introduce their own evidence, but not with industry as a sector per se.

The status of scientific evidence supplied is also meaningful for our interpretation of honest brokers in AI policymaking. Preprints made up a substantial share of cited research, and the rate of preprint citations did not differ across commenter types. Again, we see this as a means to introduce new ideas for regulation, but preprint citation alone does not distinguish broadening from narrowing of policy choices alone. Whether preprints broaden the options before regulators depends on whether they represent a range of findings or a selection chosen to support a position. 

\section{Limitations}
The primary limitation of this work is that we do not examine the context in which a citation is used, so we cannot say what \textit{effect} commenters intended a citation to have. We can therefore only provide descriptive observations about who cited what kinds of evidence. A second limitation is that websites are the most frequently cited source category, but we do not know what content these websites contain. Websites can host a wide range of material, from technical reports and company blog posts to non-academic articles discussing the social and technical harms of AI. We attempt to mitigate this by providing a more granular breakdown of the sources classified as websites. Third, we do not know the effect that these citations have on regulators. Because many of the proposed rules or requests for information have not resulted in actual regulation, an outcome variable for us to test influence is not available. 

\section{Conclusion}
This paper provides a quantitative description of the evidence introduced into federal AI policymaking and by whom. We find evidence that commenters differ significantly on the amount and type of evidence that they introduce to potential AI regulation. But, despite organizational sorting or even evidence preferences, we see many examples of honest brokers that are highly engaged with and participating in the rulemaking process for potential regulation. Future work might operationalize these concepts more cleanly, for example, by testing how the rhetorical force in a citation sentence differs by evidences type. We have also largely focused on the supply side of evidence, and as Wei et al.~\cite{wei2024how} have shown, further investigating the demand side of regulation might also prove critical to understanding how honest brokers might better inform policymaking and combat information capture.

\section{Generative AI Usage Statement}
We used Gemini 2.5 Flash, and the Claude family of models to assist with data collection, data processing, statistical analysis, and revising figures (disclosed in detail throughout). We used Claude Opus 5.5 for grammatical editing, and revising sentences for clarity.

\section*{Ethical Considerations Statement}
Our study uses publicly available data and did not consist of any human subject interactions. We identify a number of public entities in our dataset, but do not disclose or release data that would allow for this publicly identifiable information to be accessed. We identify organizations by name in our work, but we believe this is both justified and responsible given the public nature of public comments.  

\clearpage

\bibliographystyle{ACM-Reference-Format}
\bibliography{sample-base_1}

\clearpage

\section{Appendix}
\label{sec:Appendix}

\lstset{
    frame=single,               
    basicstyle=\ttfamily,       
    keywordstyle=\color{blue},  
    commentstyle=\color{gray},  
    numbers=left,               
    breaklines=true             
}

\subsubsection{Data Collection.}
\label{sec:app_data_collection}

We collect all federal dockets that may be relevant to artificial intelligence regulation by querying the GovInfo’s Federal Register API with the following search terms: "Artificial Intelligence", “AI”, "Frontier Model", "Open Weight Model", "Artificial General Intelligence", "General Purpose AI", “Large Language Model”, and "Open Model Weights". This query returned 208 results.

\subsubsection{Filtering Public Comments.}
\label{sec:app_filtering_public_comments}

We remove all comments that are less than 2 pages, as prior work has identified that it is unlikely that these comments contain substantive content \cite{costa2019public}. Of the 10,452 comments excluded for being shorter than two pages, 8,327 (79.67\%) were authored by individuals, who prior work has shown tend to submit low-information comments~\cite{dinour2017potato}. We do not have commenter-type metadata for the remaining 2,125 excluded comments. Most of these (2,024; 95.2\%) are form comments, which are short (mean = 183.7 words; median = 113 words).

Moreover, we remove all secondary attachments within a submission. As a single submission can contain several attachments, we consider all attachments beyond the first one to be secondary attachments. We manually reviewed all 165 submissions that contained several attachments and found that attachments beyond the first are predominantly duplicated comments in a different file type or supplementary materials, including research papers, news articles, and images, as opposed to text written by the commenter. We exclude supplementary materials because we are interested in the citations commenters make directly. 

\subsubsection{Public Comment Deduplication} 
\label{sec:app_comment_deduplication}

\begin{figure}[htbp]
    \centering
    \includegraphics[width=1.0\textwidth]{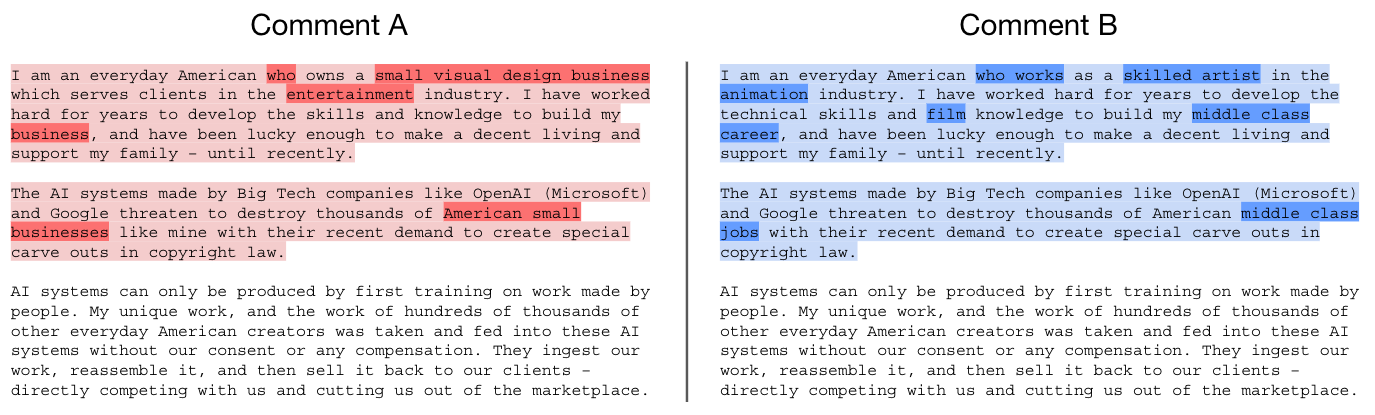}
    \caption{Identified Comment Duplicates}
    \label{fig:comment_dedup}
\end{figure}

We do not include in our analysis comments that are exact- or near-duplicates. We define exact duplicate comments as verbatim, copy-paste comments, which often come from mass comment campaigns. “Near-duplicate” comments are defined as comments that contain nearly identical body-text with the exception of non-substantive modifications. Removing these comments does not affect the results of this study, as duplicate comments contribute no additional analytical information beyond what is already represented in the corpus. 

First, we process the markdown text by stripping whitespace from the text, lowercasing all text, and removing markdown syntax, including removing hashes and asterisks. We identify duplicate comments in two passes, applied within each docket: (1) exact duplicates are identified using exact-string matching on the full text; (2) near-duplicates are identified by computing normalized Indel similarity for all possible pairs of comments within a docket. All pairs with a normalized Indel similarity above 0.90 are flagged as near-duplicates. Indel similarity – which captures surface-level textual similarity –  is the appropriate metric to use for this task because our objective is to remove verbatim or near-verbatim duplicate comments, as opposed to those that are semantically similar. 


We manually reviewed 40 identified exact-duplicate pairs and all 47 near-duplicate pairs. Manual review confirmed that a threshold of 90\% is appropriate, as it captures document pairs whose differences are non-substantive (e.g., variation in commenter metadata or the removal of a small number of sentences). See Figure~\ref{fig:comment_dedup}.

\subsubsection{Comment Text Cleaning} 
\label{sec:app_clean_md}

To measure the substantive content each commenter provided, we cleaned the comment text before computing word counts. We removed (1) citation links; (2) footnotes and reference lists; (3) submission metadata and RFI titles appended to the top of the comment; (4) required disclosure statements (e.g., ``This comment is for public disclosure...’’); (5) questions from the Request for Information (RFI) that commenters restated before answering; and (6) signature blocks and lists of signatories.

\subsubsection{Public Comment Data Processing Pipeline}

\subsubsection{Prompts}

\label{sec:app_prompts}


Prompt used for comments containing hyperlinks:
\begin{Verbatim}[breaklines=true, breakanywhere=false, fontsize=\footnotesize, frame=single]
You are performing STRUCTURAL IDENTIFICATION AND EXTRACTION on a document that has been converted to Markdown.
The document contains citations in the form of in-text hyperlinks. Your task is to identify, label, and extract the in-text hyperlinks and other structural elements from the document. Do NOT delete content. If an element does not belong to the main body, classify and return it separately.
---
## Category 1 - Citations
This is a deterministic information extraction task. Extract ALL citations from the document. This must be exhaustive.
### What IS a Citation Sentence
A citation sentence is body text that contains an EXPLICIT citation marker, such as:
- Superscript or bracketed footnote numbers: [1], [2], <sup>1</sup>, <sup>2</sup>, etc.
- Bracketed reference numbers: [Smith 2020], [12], etc.
- Inline hyperlinks to external sources (NOT submission portals or contact info)
Return citation-sentence and in-text hyperlink pairs. These pairs are often formatted as (citation_sentence)[citation_link]. 
1. **citation_sentence**: The sentence(s) in the body paragraph containing the citation marker. Include only the sentence(s) directly containing or immediately surrounding the marker.
2. **citation_link**: The inline hyperlink associated with that citation marker.
### What is NOT a Citation Sentence (EXCLUDE these)
- **Submission/introduction text**: Headers like "BY ELECTRONIC SUBMISSION VIA...", "RE: Docket No...", "Subject:", "Submitted By:", document titles
- **Docket/reference numbers in headers**: e.g., "(DOCKET NO. FDA-2024-D-4689)"
- **Quoted text without citation markers**: Text in quotation marks is NOT automatically a citation -- it requires an explicit marker
- **Contact information**: Email addresses, LinkedIn profiles, author bios
- **Footnote text itself**: The footnote is the REFERENCE, not the citation sentence
- **Mere mentions of external documents**: Mentioning "the 21st Century Cures Act" or "GDPR" or "AI RMF 2.0" is NOT a citation unless accompanied by an explicit citation marker
- **Submission portal URLs**: regulations.gov, mailto: links for document submission
### For Other Citation Types Present
If non-reference citations exist, return:
1. **citation_sentence**: The body text containing the citation marker
2. **citation_reference**: The reference list entry (if present), including hierarchical numbers
3. **citation_footnotes**: The footnote text associated with the citation sentence (if present)
NOTE: A citation_sentence may contain multiple citation markers. In that case, return one pair per citation marker.  Repeating the full citation_sentence in each pair, but with only the corresponding single citation_reference. DO NOT combine multiple references into one citation_reference entry.
### Handling Edge Cases
- **Hyperlinks spanning multiple sentences**: Return EACH sentence separately, each paired with the same link
- **Duplicate links**: Return ALL instances, even if the same URL appears multiple times
- **DOIs**: Treat `doi:` or `DOI:` prefixed strings as citation links (e.g., "doi: 10.1055/s-0038-1625964")
- **Links in footnotes**: Extract URLs from within footnote text and include them in citation_links
### Rules
1. Do not summarize, rank, filter, or deduplicate
2. Preserve original text exactly (including line breaks as they appear)
3. A citation REQUIRES an explicit marker -- do not infer citations from context
---
## Category 3 - Authorship
Return:
1. Author name(s)
2. Author affiliations
3. Author contact info
**Note**: Author contact info belongs HERE, not in citations.
---
## Category 4 - Reference List / Bibliography
Return:
1. The entire reference list or bibliography, typically located at the end of the document
2. Do NOT compile footnotes to create a bibliography -- return only an explicitly labeled reference list if one exists
---
## OUTPUT FORMAT
Return the extracted text verbatim from the document for each field.
You must strictly output JSON adhering to the following schema:
\end{Verbatim}

Prompt used for comments containing footnotes:
\begin{Verbatim}[breaklines=true, breakanywhere=false, fontsize=\footnotesize, frame=single]
You are performing STRUCTURAL IDENTIFICATION AND EXTRACTION on a document that has been converted to Markdown.
The document contains citations in the form of footnotes. Your task is to identify, label, and extract the footnotes and other structural elements from the document. Do NOT delete content. If an element does not belong to the main body, classify and return it separately.
---
## Category 1 - Citations
This is a deterministic information extraction task. Extract ALL citations from the document. This must be exhaustive.
### What IS a Citation Sentence
A citation sentence is body text that contains an EXPLICIT citation marker, such as:
- Superscript or bracketed footnote numbers: [1], [2], <sup>1</sup>, <sup>2</sup>, etc.
- Bracketed reference numbers: [Smith 2020], [12], etc.
- Inline hyperlinks to external sources (NOT submission portals or contact info)
Return citation-sentence and footnote pairs:
1. **citation_sentence**: The sentence(s) in the body paragraph containing the citation marker. Include only the sentence(s) directly containing or immediately surrounding the marker.
2. **citation_footnote**: The footnote text associated with that citation marker.
### What is NOT a Citation Sentence (EXCLUDE these)
- **Submission/introduction text**: Headers like "BY ELECTRONIC SUBMISSION VIA...", "RE: Docket No...", "Subject:", "Submitted By:", document titles
- **Docket/reference numbers in headers**: e.g., "(DOCKET NO. FDA-2024-D-4689)"
- **Quoted text without citation markers**: Text in quotation marks is NOT automatically a citation -- it requires an explicit marker
- **Contact information**: Email addresses, LinkedIn profiles, author bios
- **Footnote text itself**: The footnote is the REFERENCE, not the citation sentence
- **Mere mentions of external documents**: Mentioning "the 21st Century Cures Act" or "GDPR" or "AI RMF 2.0" is NOT a citation unless accompanied by an explicit citation marker
- **Submission portal URLs**: regulations.gov, mailto: links for document submission
### For Other Citation Types Present
If non-footnote citations exist, return:
1. **citation_sentence**: The body text containing the citation marker
2. **citation_link**: The in-text hyperlink (if present)
3. **citation_reference**: The reference list entry (if present), including hierarchical numbers
NOTE: A citation_sentence may contain multiple citation markers. In that case, return one pair per citation marker — repeating the full citation_sentence in each pair, but with only the corresponding single citation_reference. DO NOT combine multiple references into one citation_reference entry.
### Handling Edge Cases
- **Hyperlinks spanning multiple sentences**: Return EACH sentence separately, each paired with the same link
- **Duplicate links**: Return ALL instances, even if the same URL appears multiple times
- **DOIs**: Treat `doi:` or `DOI:` prefixed strings as citation links (e.g., "doi: 10.1055/s-0038-1625964")
- **Links in footnotes**: Extract URLs from within footnote text and include them in citation_links
### Rules
1. Do not summarize, rank, filter, or deduplicate
2. Preserve original text exactly (including line breaks as they appear)
3. A citation REQUIRES an explicit marker -- do not infer citations from context
---
## Category 3 - Authorship
Return:
1. Author name(s)
2. Author affiliations
3. Author contact info
**Note**: Author contact info belongs HERE, not in citations.
---
## Category 4 - Reference List / Bibliography
Return:
1. The entire reference list or bibliography, typically located at the end of the document
2. Do NOT compile footnotes to create a bibliography -- return only an explicitly labeled reference list if one exists
## OUTPUT FORMAT
Return the extracted text verbatim from the document for each field.
You must strictly output JSON adhering to the following schema:
\end{Verbatim}

Prompt used for comments containing a bibliography/reference list:
\begin{Verbatim}[breaklines=true, breakanywhere=false, fontsize=\footnotesize, frame=single]
You are performing STRUCTURAL IDENTIFICATION AND EXTRACTION on a document that has been converted to Markdown.
The document contains citations in the form of bibliographic references. Your task is to identify, label, and extract the bibliographic references and other structural elements from the document. Do NOT delete content. If an element does not belong to the main body, classify and return it separately.
---
## Category 1 - Citations
This is a deterministic information extraction task. Extract ALL citations from the document. This must be exhaustive.
### What IS a Citation Sentence
A citation sentence is body text that contains an EXPLICIT citation marker, such as:
- Superscript or bracketed footnote numbers: [1], [2], <sup>1</sup>, <sup>2</sup>, etc.
- Bracketed reference numbers: [Smith 2020], [12], etc.
- Author-year citations: (Smith, 2020), (Smith et al. 2020; Lee 2019), Smith (2020), etc.
- Roman numeral endnote markers: <sup>iv</sup>, [xii], etc.
- Inline hyperlinks to external sources (NOT submission portals or contact info)
Return citation-sentence and bibliographic entry pairs:
1. **citation_sentence**: The sentence(s) in the body paragraph containing the citation marker. Include only the sentence(s) directly containing or immediately surrounding the marker.
2. **citation_reference**: The full bibliographic entry from the reference list/bibliography that corresponds to this citation marker.
NOTE: A citation_sentence may contain multiple citation markers. In that case, return one pair per citation marker — repeating the full citation_sentence in each pair, but with only the corresponding single citation_reference. DO NOT combine multiple references into one citation_reference entry.
### Matching In-Text Citations to Reference Entries
The reference list may be far from the citation sentence (e.g., at the end of the document, or appended after a final '---' separator). For EVERY citation marker, look up the matching entry in the reference list and copy that entry verbatim into citation_reference. Do NOT leave citation_reference empty when a matching entry exists. Match using the signal the marker carries:
- **Author-year markers** — e.g., "(Dan, 2021)", "Dan et al. (2021)", "(Dan and Lee 2021; Smith 2019)": match the entry whose first author's surname starts the entry (e.g., the entry beginning "Dan, ..." or "Dan, J., ...") and whose year matches. Use the year (and suffixes like 2021a / 2021b, or co-author names) to choose between entries by the same first author.
- **Numeric markers** — e.g., [3], [3, 7], [3-5], <sup>3</sup>, or a bare trailing number: match the entry numbered 3 (e.g., "3.", "[3]", "3 " at the start of the entry). Expand ranges ("[3-5]" -> 3, 4 and 5) and lists ("[3, 7]" -> 3 and 7).
- **Roman numeral markers** — e.g., <sup>iv</sup>, [xii], or "xxxii": match the entry that begins with that numeral (e.g., "iv", "xii.", "xxxii").
- **Other labels** — e.g., [Smith20], [OECD]: match the entry that begins with or is labeled by the same key.
When one sentence cites several entries — e.g., "(Dan, 2021; Lee & Park, 2019)" or "[2, 5, 9]" — return a SEPARATE pair for EACH cited entry, each with the full citation_sentence and that single entry as citation_reference. If, after checking the whole reference list, no entry matches a marker, still return the citation_sentence with citation_reference set to null.
### What is NOT a Citation Sentence (EXCLUDE these)
- **Submission/introduction text**: Headers like "BY ELECTRONIC SUBMISSION VIA...", "RE: Docket No...", "Subject:", "Submitted By:", document titles
- **Docket/reference numbers in headers**: e.g., "(DOCKET NO. FDA-2024-D-4689)"
- **Quoted text without citation markers**: Text in quotation marks is NOT automatically a citation -- it requires an explicit marker
- **Contact information**: Email addresses, LinkedIn profiles, author bios
- **Footnote text itself**: The footnote is the REFERENCE, not the citation sentence
- **Mere mentions of external documents**: Mentioning "the 21st Century Cures Act" or "GDPR" or "AI RMF 2.0" is NOT a citation unless accompanied by an explicit citation marker
- **Submission portal URLs**: regulations.gov, mailto: links for document submission
### For Other Citation Types Present
If non-reference citations exist, return:
1. **citation_sentence**: The body text containing the citation marker
2. **citation_link**: The in-text hyperlink (if present)
3. **citation_footnotes**: The footnote text associated with the citation sentence (if present)
### Handling Edge Cases
- **Hyperlinks spanning multiple sentences**: Return EACH sentence separately, each paired with the same link
- **Duplicate links**: Return ALL instances, even if the same URL appears multiple times
- **DOIs**: Treat `doi:` or `DOI:` prefixed strings as citation links (e.g., "doi: 10.1055/s-0038-1625964")
- **Links in footnotes**: Extract URLs from within footnote text and include them in citation_links
### Rules
1. Do not summarize, rank, filter, or deduplicate
2. Preserve original text exactly (including line breaks as they appear)
3. A citation REQUIRES an explicit marker -- do not infer citations from context
---
## Category 3 - Authorship
Return:
1. Author name(s)
2. Author affiliations
3. Author contact info
**Note**: Author contact info belongs HERE, not in citations.
---
## Category 4 - Reference List / Bibliography
Return:
1. The entire reference list or bibliography, typically located at the end of the document
2. Do NOT compile footnotes to create a bibliography -- return only an explicitly labeled reference list if one exists
---
## OUTPUT FORMAT
Return the extracted text verbatim from the document for each field.
You must strictly output JSON adhering to the following schema:
\end{Verbatim}

\subsubsection{Model Evaluation}
\label{sec:app_model_eval}

\subsubsection{Post-Processing Steps}
\label{sec:app_post_processing}

We process both the human and model annotations by removing all whitespace, removing all Markdown and HTML syntax, and removing all hyperlinked text found in the citation sentences and citation footnotes/references (e.g., “This text is [hyperlinked](url)” becomes “This text is hyperlinked”). We choose to remove the links because markdown conversion of PDFs frequently returns broken, fragmented links due to line breaks. 

We manually review the model’s extractions, and find that the model may incorrectly return mailto links as citation links or return fragmented citation footnotes/references (e.g., Author, Year is returned, but not the remaining citation information). We remove these from the model’s annotations. We also find that when a single citation sentence contains multiple bibliographic references, the model returns them grouped as a single citation reference. We parse and split these into individual references. Additionally, we observe that the model may not extract citation links that are found in citation footnotes/references. We identify and extract links found in the citation footnotes/references using URLExtract, and then append these links as individual citation links. Lastly, we initially distinguished between in-text citations and footnote citations. However, the model frequently mislabeled which type an extracted citation was, even though it still correctly extracted the sentence–citation pair. We therefore chose to combine these categories.

\subsubsection{Evaluation Method}
\label{sec:app_eval}

\begin{table}[htbp]
\centering
\small
\caption{Human vs.\ Model Annotations}
\label{tab:human-vs-model}
\renewcommand{\arraystretch}{1.3}
\begin{tabularx}{\textwidth}{@{} >{\raggedright\arraybackslash}p{2.6cm}
                                 >{\raggedright\arraybackslash}X
                                 >{\raggedright\arraybackslash}X @{}}
\toprule
\textbf{Label} & \textbf{Ground Truth} & \textbf{Model's Annotation} \\
\midrule
\texttt{citation\_sentence} &
A dynamic, risk-based approach to AI regulation is necessary (National Institute of Standards and Technology [NIST], 2023). &
I. Dynamic, Risk-Based Approach: A dynamic, risk-based approach to AI regulation is necessary (National Institute of Standards and Technology [NIST], 2023). \\
\addlinespace
\texttt{citation\_link} &
\url{https://nvlpubs.nist.gov/nistpubs/ai/NIST.AI.100-1.pdf} &
\url{https://nvlpubs.nist.gov/nistpubs/ai/NIST.AI.100-1.pdf} \\
\addlinespace
\texttt{citation\_footnote/ reference} &
National Institute of Standards and Technology. (2023). \textit{Artificial intelligence risk management framework (AI RMF 1.0)}. \url{https://nvlpubs.nist.gov/nistpubs/ai/NIST.AI.100-1.pdf} &
National Institute of Standards and Technology. (2023). \textit{Artificial intelligence risk management framework (AI RMF 1.0)}. \url{https://nvlpubs.nist.gov/nistpubs/ai/NIST.AI.100-1.pdf} \\
\bottomrule
\end{tabularx}

\vspace{4pt}
\begin{minipage}{\textwidth}
\footnotesize
\textit{Note.} In this example, \texttt{citation\_sentence} receives a normalized Levenshtein similarity score of 0.788, the lowest score in our dataset. The model's annotation includes additional context (``I. Dynamic, Risk-Based Approach:'') that the human annotation does not. This addition neither alters the meaning of the citation sentence nor affects the extraction of the cited source. \texttt{citation\_link} and \texttt{citation\_footnote/reference} each achieve a similarity score of 1, indicating an exact match.
\end{minipage}
\end{table}

Following Breton et al.~\cite{breton2025leveraging}, we evaluate the model’s citation information extraction – including citation sentences, citation links, and citation footnotes/references – against the ground truth dataset by using two matching criteria: (1) exact-string matches and (2) near-string matches via normalized Levenshtein similarity. Normalized Levenshtein similarity measures the similarity between two text strings by measuring the minimum number of insertions, deletions, and substitutions needed to transform one into the other~\cite{levenshtein1966binary}. Moreover, we observe that the model frequently returned both the author's affiliation and job title. We re-process these affiliation strings by using gemini-2.5-flash to extract just the affiliation. 

We treat near-matches or partial matches as valid matches for three reasons. First, minor differences in syntactic markers (e.g., [16]) can produce false non-matches. Second, the model may include surrounding text or truncate the citation span, but the cited information is still present. Lastly, this does not affect the extraction of the cited information, which is the primary unit of analysis. See Table~\ref{tab:human-vs-model}. 

For citation sentences and citation footnotes/references, we consider only ground truth-model annotation pairs (ground truth citation sentence vs. model extracted citation sentence; ground truth citation footnote/reference vs. model extracted citation footnote/reference) that score above 0.75 to be near-string matches. For citation links, we consider scores greater than 0.80 to be near matches, as PDF conversion to markdown may result in garbled links. For citation sentences only, we consider partial sub-string overlap to be valid matches. We recognize that span boundary similarity (as done for IRR), is the ideal method of evaluating the model’s span extractions. However, the model does not return nor do we prompt for the model to return the span’s location in the markdown document. We attempted to identify the location of the spans in the markdown document by overlaying the extracted spans over the document, but were unable to develop a reliable method to do so. Relying on string matching and string overlap, as done by Breton et al.~\cite{breton2025leveraging}, is the most plausible alternative. 

\subsubsection{Classifying Commenter Type}
\label{sec:app_commenter_type}
Comments are submitted either by individuals or on behalf of organizations. We obtain author names and organizational affiliations from the metadata returned by the Regulations.gov API; where these fields are missing, we extract them from the comment text. To characterize organizational commenters, we assign each affiliation a North American Industry Classification System (NAICS) code. NAICS is the federal standard for classifying establishments by their primary economic activity, with 1,012 six-digit industries in its 2022 version.

We developed a Claude skill assigns each organization a NAICS codes When an organization lacks a description, the model searches the web and writes a three-sentence summary of its activities. A keyword-weighted shortlist narrows the candidate codes, and the model selects the code that matches the organization’s primary activity. A script then validates every code and title against the official 2022 Census list. Each record receives a confidence score (0 to 1) and a written justification. Records scored below 0.95 also receive three ranked alternate codes.

Commenters who participate in several dockets often report their affiliation differently each time (e.g., Google LLC vs. Google Inc.) or give the name of a subsidiary. When research shows an entity is a subsidiary, division, or program of a larger organization, the skill records the parent-qualified name and codes the entity by the parent’s activity. On the validation set of 100, the skill achieved 84\% accuracy. We retain organizations classified with confidence above 0.9 ($n$ = 953 of 1,582). We then map each NAICS code to one of five organizational categories: civil society organization, trade/coalition organization, business/industry, university, or government. Individuals form the sixth category.


Because the same organization often appears under different names across dockets (e.g., Google LLC, Google Inc., or a subsidiary such as DeepMind), matching affiliation strings directly to industry codes would split one organization into several entities and code them inconsistently. To resolve these differences, we use a Claude skill that researches each organization name and selects a code from the official 2022 Census list. We retain codes assigned with confidence above 0.9 and map each one to an organizational category. Following this process, 54 organizations were unclassified. We manually classify these organizations.

\subsubsection{Classifying Evidence Type.}
\label{sec:app_evi_type}


We classify each cited source into one of four evidence categories: research article, legal citation, government resource, or website. We further divide websites into news articles, educational resources, and social media posts. Table~\ref{tab:evidence-detection} outlines the rule-based detection method and the post-processing steps used for each evidence type. For the remaining citations that the rule-based method could not classify ($n$ = 2,714), we programmatically classify them using a Claude skill. For each citation, the skill searches the web using the reference text, identifies the most likely source, and assigns it to one of the evidence categories above. It also returns structured metadata for the identified source, including title, author(s), and publication year. To check that the skill found the correct source, we compare the returned metadata against the fields extracted from the original citation. Sources whose metadata match the citation receive a high confidence score; partial matches receive a medium score, and sources with no match receive a low score. We retain only the citations that received a high confidence score. 

Of the 2,714 citations processed by the skill, 1,355 (49.93\%) received a high confidence score and were retained. Most of the remaining citations ($n$ = 752) were not references to external evidence but footnote text, such as explanatory notes or commentary. We excluded these from our analysis.

\begin{table*}[t]
\centering
\small
\setlength{\tabcolsep}{4pt}
\renewcommand{\arraystretch}{1.2}
\begin{tabular}{>{\raggedright\arraybackslash}p{0.17\textwidth}
                >{\raggedright\arraybackslash}p{0.22\textwidth}
                >{\raggedright\arraybackslash}p{0.25\textwidth}
                >{\centering\arraybackslash}p{0.08\textwidth}
                >{\centering\arraybackslash}p{0.08\textwidth}
                >{\centering\arraybackslash}p{0.07\textwidth}}
\toprule
\textbf{Type of Evidence} & \textbf{Rule-Based Detection} & \textbf{Post-Processing and Validation} & \textbf{N (Rule-Based)} & \textbf{N (LLM-Based)} & \textbf{Total} \\
\midrule
Research Article &
  URL matches scholarly domains (e.g., \texttt{doi.org}, \texttt{arxiv.org}, \texttt{ssrn.com}) &
  Extract DOI or arXiv ID; otherwise confirm via CrossRef or OpenAlex &
  5,276 & 88 & 5,364 \\
Legal Citation &
  \texttt{eyecite}~\cite{cushman2021eyecite} and regex patterns for U.S.C.\ and C.F.R.\ citations &
  Remove references to E.O.~14110, E.O.~14179, and RFI numbers &
  1,446 & 232 & 1,678 \\
Government Resource &
  Government top-level domain (\texttt{.gov}, \texttt{.mil}) &
  -- &
  4,845 & 250 & 5,095 \\
Website & & &
  13,830 & 573 & 14,403 \\
\quad $\hookrightarrow$ News Article &
  Domain in \cite{lin2023high} dataset of 11,520 news domains &
  -- &
  3,390 & 107 & 3,497 \\
\quad $\hookrightarrow$ Edu. Resource &
  Academic top-level domain (\texttt{.edu}) &
  Extract institution name using \texttt{urlparse} and \texttt{tldextract} &
  914 & -- & 914 \\
\quad $\hookrightarrow$ Social Media &
  Domain in platform list &
  -- &
  276 & -- & 276 \\
\quad $\hookrightarrow$ Other &
  Remaining citation links &
  Extract domains using \texttt{urlparse} and \texttt{tldextract} &
  9,250 & 466 & 9,716 \\
\midrule
\textbf{Total} & & & \textbf{25,397} & \textbf{1,143} & \textbf{26,540} \\
\bottomrule
\end{tabular}
\caption{Rule-based detection and post-processing steps by type of evidence}
\label{tab:evidence-detection}
\end{table*}

\subsubsection{Retrieving Bibliographic Data of Cited Research Articles.}
\label{sec:app_bib_data}

For the research articles for which we were unable to retrieve a DOI from Crossref with high confidence (>90\%), we use few-shot prompting with \texttt{gemini-2.5-flash} to parse each citation’s reference text for papers' title, author(s), venue, and publication year. We then query the OpenAlex API with the parsed title and verify that the returned record is a correct match by comparing it against the parsed author names, year, and/or venue. Moreover, we found that querying OpenAlex with an arXiv ID frequently returned an incorrect record. To address this, we additionally query the arXiv API to retrieve each paper’s title, authors, and year, and then match these against the OpenAlex record. We manually review 161 citation references, and find that we are able to return 87.58\% (141/161) correct OpenAlex records.

Lastly, we supplement the OpenAlex data to determine whether each cited paper was a preprint or a peer-reviewed article at the time the comment was submitted. Relying on the OpenAlex data alone, we observe numerous discrepancies in OpenAlex's publication status and the version of the article that the commenter cites. For example, the commenter cites the peer-reviewed version of a paper, but OpenAlex lists the paper as preprint, or vice versa. To resolve these discrepancies, we use the Semantic Scholar API to retrieve the date that a preprint was posted to arXiv and, if it was later-peer-reviewed, the date of its peer-reviewed publication. We then apply two rules: First, if a comment was submitted before the peer-reviewed publication date, we treat the comment as citing a preprint. Second, if OpenAlex lists a work as a preprint but the commenter’s reference names a journal or conference, we treat the work as peer-reviewed. Mapping together these publication and submission dates, we identify the works that were cited in comments as preprints and later became peer-reviewed articles. To validate the coding of publication status, we manually reviewed a random sample of 180 papers. Of these, 169 (93.89\%) were coded correctly.

\subsection{Identifying Self-Citation}
\label{sec:ident_self_cit}

As noted above, we define self-citation as a citation to evidence produced by the commenter's own organization. Self-citation can take two forms: a citation to the organization's own website (website self-citation) or to a research paper written by authors affiliated with the organization (academic self-citation).

Both forms require resolving each commenter organization to its web domains and Research Organization Registry (ROR) identifiers. We draw these from three sources: (1) records in the ROR database whose name exactly matches the organization's name; (2) Wikidata entries whose name exactly matches and that list a website or ROR identifier; and (3) a manually compiled list of 308 organization domains.

To identify website self-citations, we clean each cited URL and extract its domain using \cite{urlextract}. A citation is a website self-citation when that domain matches one of the commenter's domains.

To identify academic self-citations, we take the institutional affiliations of each cited paper's authors, with their ROR identifiers, from OpenAlex. A citation is an academic self-citation when one of these institutions shares a ROR identifier or web domain with the commenter. When an institution has no ROR identifier, its name must exactly match the commenter's name.

\subsection{Results}

\begin{table*}[t]
\centering
\small
\caption{Negative binomial mixed model of citation counts by evidence type and commenter type.}
\label{tab:rq1_evidence_by_commenter}
\begin{tabular}{l r r r r l l}
\toprule
\textbf{Term} & \textbf{Est.} & \textbf{SE} & \textbf{z} & \textbf{IRR} & \textbf{95\% CI} & \textbf{p} \\
\midrule
\multicolumn{7}{l}{\textit{Evidence type (ref: Website)}} \\
\rowcolor{gray!30}
\quad Research article        & $-1.762$ & 0.103 & $-17.06$ & 0.17 & [0.14, 0.21] & $<$.001 \\
\rowcolor{gray!30}
\quad Government resource     & $-1.126$ & 0.095 & $-11.85$ & 0.32 & [0.27, 0.39] & $<$.001 \\
\rowcolor{gray!30}
\quad Legal citation          & $-2.824$ & 0.124 & $-22.69$ & 0.06 & [0.05, 0.08] & $<$.001 \\
\addlinespace
\multicolumn{7}{l}{\textit{Commenter type (ref: Business/Industry)}} \\
\rowcolor{gray!30}
\quad Individual                 & $-0.804$ & 0.160 & $-5.02$ & 0.45 & [0.33, 0.61] & $<$.001 \\
\quad Government                 & 0.473    & 0.359 & 1.32    & 1.61 & [0.79, 3.25] & .188 \\
\rowcolor{gray!30}
\quad Civil society organization & 0.923    & 0.157 & 5.86    & 2.52 & [1.85, 3.43] & $<$.001 \\
\rowcolor{gray!30}
\quad Trade/coalition            & 0.516    & 0.152 & 3.39    & 1.68 & [1.24, 2.26] & $<$.001 \\
\quad University                 & 0.348    & 0.219 & 1.59    & 1.42 & [0.92, 2.18] & .113 \\
\addlinespace
\rowcolor{gray!30}
Log word count (centered)     & 1.247    & 0.047 & 26.82   & 3.48 & [3.18, 3.81] & $<$.001 \\
\addlinespace
\multicolumn{7}{l}{\textit{Interactions}} \\
\quad Individual $\times$ & & & & & & \\
\qquad Government resource & $-0.277$ & 0.172 & $-1.61$ & 0.76 & [0.54, 1.06] & .108 \\
\qquad Legal citation      & 0.047    & 0.231 & 0.20    & 1.05 & [0.67, 1.65] & .838 \\
\qquad Research article    & 0.305    & 0.178 & 1.71    & 1.36 & [0.96, 1.92] & .087 \\
\quad Government $\times$ & & & & & & \\
\rowcolor{gray!30}
\qquad Government resource & 0.731    & 0.338 & 2.16    & 2.08 & [1.07, 4.03] & .030 \\
\qquad Legal citation      & 0.548    & 0.411 & 1.33    & 1.73 & [0.77, 3.87] & .183 \\
\rowcolor{gray!30}
\qquad Research article    & 1.243    & 0.344 & 3.62    & 3.46 & [1.77, 6.80] & $<$.001 \\
\quad Civil society organization $\times$ & & & & & & \\
\qquad Government resource & $-0.208$ & 0.139 & $-1.50$ & 0.81 & [0.62, 1.07] & .134 \\
\qquad Legal citation      & $-0.117$ & 0.175 & $-0.67$ & 0.89 & [0.63, 1.25] & .504 \\
\rowcolor{gray!30}
\qquad Research article    & 0.320    & 0.148 & 2.17    & 1.38 & [1.03, 1.84] & .030 \\
\quad Trade/coalition $\times$ & & & & & & \\
\qquad Government resource & 0.241    & 0.134 & 1.79    & 1.27 & [0.98, 1.65] & .073 \\
\rowcolor{gray!30}
\qquad Legal citation      & 0.600    & 0.166 & 3.62    & 1.82 & [1.32, 2.52] & $<$.001 \\
\rowcolor{gray!30}
\qquad Research article    & $-0.378$ & 0.150 & $-2.52$ & 0.68 & [0.51, 0.92] & .012 \\
\quad University $\times$ & & & & & & \\
\qquad Government resource & 0.019    & 0.187 & 0.10    & 1.02 & [0.71, 1.47] & .919 \\
\qquad Legal citation      & $-0.294$ & 0.258 & $-1.14$ & 0.75 & [0.45, 1.24] & .254 \\
\rowcolor{gray!30}
\qquad Research article    & 1.711    & 0.189 & 9.05    & 5.54 & [3.82, 8.02] & $<$.001 \\
\addlinespace
Intercept        & 0.376    & 0.123 & 3.05   & 1.46 & [1.14, 1.85] & .002 \\
\midrule
\multicolumn{7}{l}{\textit{Random effects (variance)}} \\
\quad Commenter  & \multicolumn{6}{l}{1.932 (SD = 1.390; 1,799 commenters)} \\
\quad Comment    & \multicolumn{6}{l}{0.521 (SD = 0.722; 2,664 comments)} \\
\quad Docket     & \multicolumn{6}{l}{0.069 (SD = 0.262; 26 dockets)} \\
NB dispersion ($\phi$) & \multicolumn{6}{l}{0.670} \\
\midrule
Observations     & \multicolumn{6}{l}{10,656} \\
\bottomrule
\end{tabular}
\vspace{2pt}
\begin{minipage}{0.95\textwidth}
\footnotesize
\textit{Note.} This table contains the results from the negative binomial model described in Section~\ref{sec:Evidence_use_by_Commenter_Type}, which models the relationship between \textit{Commenter Type} and \textit{Evidence Type}.
\end{minipage}
\end{table*}

\begin{table}[htbp]
\centering
\caption{Negative Binomial of self-citation rates}
\label{tab:nb_self_citation}
\begin{tabular}{lrrrrcr}
\toprule
 & Estimate & SE & $z$ & IRR & 95\% CI & $p$ \\
\midrule
(Intercept)              & 0.003 & 0.127 & 0.02 & 1.003 & [0.782, 1.287] & 0.980 \\
\quad Government                 & 0.357 & 0.274 & 1.30 & 1.429 & [0.83, 2.44] & 0.193 \\
\quad Civil society org.         & $-0.015$ & 0.139 & $-0.11$ & 0.985 & [0.75, 1.29] & 0.915 \\
\quad Trade/coalition org.       & $-0.008$ & 0.137 & $-0.06$ & 0.992 & [0.76, 1.30] & 0.951 \\
\quad University                 & 0.296 & 0.183 & 1.62 & 1.344 & [0.94, 1.92] & 0.106 \\
\addlinespace
log(word count), centered        & $-0.056$ & 0.062 & $-0.92$ & 0.945 & [0.84, 1.07] & 0.360 \\
log(N evaluable, any family), centered & 0.856 & 0.046 & 18.65 & 2.353 & [2.15, 2.57] & $<0.001$ \\
\bottomrule
\end{tabular}
\vspace{2pt}
\begin{minipage}{\linewidth}
\footnotesize
\textit{Note.} This table contains the results from the negative binomial model described in Section~\ref{sec:self-citation}, which models the relationship between \textit{Commenter Type} and the rate at which commenters self-cite.
\end{minipage}
\end{table}

\begin{table}[t]
\centering
\caption{Beta-binomial mixed model of the share of cited research articles that were preprints}
\label{tab:betabinom-peer-reviewed}
\begin{tabular}{lrrrrrr}
\toprule
Predictor & Estimate & SE & $z$ & OR & 95\% CI & $p$ \\
\midrule
(Intercept)                     & $-0.619$ & 0.291 & $-2.13$ & 0.539 & [0.30, 0.95] & 0.033 \\
Civil society org.              & 0.198    & 0.224 & 0.88    & 1.219 & [0.79, 1.89] & 0.377 \\
University                      & $-0.381$ & 0.249 & $-1.53$ & 0.683 & [0.42, 1.11] & 0.126 \\
Individual                      & $-0.165$ & 0.270 & $-0.61$ & 0.848 & [0.50, 1.44] & 0.542 \\
Trade/coalition organization    & $-0.406$ & 0.261 & $-1.55$ & 0.666 & [0.40, 1.11] & 0.121 \\
Government                      & $-0.756$ & 0.421 & $-1.79$ & 0.470 & [0.21, 1.07] & 0.073 \\
log(word count), centered       & $-0.046$ & 0.090 & $-0.50$ & 0.955 & [0.80, 1.14] & 0.614 \\
\bottomrule
\end{tabular}
\vspace{2pt}
\begin{minipage}{\linewidth}
\footnotesize
\textit{Note.} This table contains the results from the beta-binomial mixed model described in Section~\ref{sec:peer-reviewed_vs_preprints}, which models the relationsip between commenter type and the rate which they cite pre-prints vs peer-reviewed articles. Odds ratios above 1 indicate higher odds of citing preprints rather than peer-reviewed articles. Commenter-type odds ratios are relative to business/industry.
\end{minipage}
\end{table}

\end{document}